\documentclass[preprint,12pt]{elsarticle}
\usepackage{amssymb}
\usepackage{amsmath}
\usepackage{graphicx}
\usepackage{booktabs}
\usepackage[unicode]{hyperref}
\usepackage{colortbl}
\usepackage{comment}
\usepackage{tabularx}
\usepackage[ruled,vlined]{algorithm2e}
\usepackage{float}
\usepackage{hyphenat}
\usepackage{multirow}
\usepackage{array}
\usepackage{makecell}
\usepackage{arydshln}
\usepackage{times}
\usepackage[T1]{fontenc}
\usepackage{bm}
\usepackage{xcolor}
\usepackage{tcolorbox}
\definecolor{rolecolor}{RGB}{0, 0, 255}
\definecolor{correctcolor}{RGB}{0, 128, 0} 
\definecolor{errorcolor}{RGB}{255, 0, 0} 
\newenvironment{customlist}
  {\list{}{%
    \setlength{\leftmargin}{0.5em}
    \setlength{\labelwidth}{0pt}%
    \setlength{\labelsep}{0pt}%
    \setlength{\itemindent}{0pt}%
    \setlength{\parsep}{0pt}
    \setlength{\itemsep}{1\baselineskip}
  }}
  {\endlist}
\journal{Knowledge-Based Systems}
\hypersetup{
  pdfauthor={Xinchun Su, Chunxu Luo, Lipeng Ma, Yixuan Li, Weidong Yang},
  pdftitle={zhanwei},
  pdfkeywords={zhanwei}
}
\begin{document}
\begin{frontmatter}
\title{MedCritical: Enhancing Medical Reasoning in Small Language Models via Self-Collaborative Correction}
\author[Xinchun]{Xinchun Su}
\ead{22300240012@m.fudan.edu.cn}

\author[Chunxu]{Chunxu Luo}
\ead{22110240100@m.fudan.edu.cn}

\author[Yixuan]{Yixuan Li}
\ead{yxli24@m.fudan.edu.cn}

\author[Weidong]{Weidong Yang\corref{cor1}}
\ead{wdyang@fudan.edu.cn}

\author[Lipeng]{Lipeng Ma\corref{cor1}}
\ead{lpma21@m.fudan.edu.cn}

\address[Xinchun,Chunxu,Yixuan,Weidong,Lipeng]{School of Computer Science, Fudan University, Shanghai, China}

\cortext[cor1]{Corresponding authors}
\begin{abstract}
In the field of medicine, complex reasoning tasks such as clinical diagnosis, treatment planning, and medical knowledge integration pose significant challenges, where small language models  often underperform compared to large language models  like GPT-4 and Deepseek. Recent knowledge distillation-based methods aim to address these issues through teacher-guided error correction, but this LLM as judge approach remains challenging in terms of cost, time, and efficiency. To circumvent this issue, we propose a novel two-stage framework, MedCritical, which uses a small language model fine-tuned by a large teacher model to play against itself. In the first stage, we extract high-level and detailed long-chain thought templates from the teacher model to guide the student model to generate more complex reasoning thoughts. In the second stage, we introduce direct preference optimization (DPO) through model self-iteration collaboration to enhance the reasoning ability of the student model by playing against the correction trajectory of the fine-tuned model during training. This model self-learning DPO approach teaches the student model to use its own error-driven insights to consolidate its skills and knowledge to solve complex problems, and achieves comparable results to traditional knowledge distillation methods using teacher models at a lower cost. Notably, our MedCritical 7B model outperforms the Taiyi and Huatuo-o1-7B models by 3.04\% and 10.12\% respectively on the CMExam benchmark, achieving new SOTA performance among 7B-class small models. All content mentioned is open-sourced at \url{https://github.com/destinybird/SP-Huatuo}.
\end{abstract}
\end{frontmatter}
\section{Introduction}
\label{sec:introduction}
LLMs, such as GPT-4 and DeepSeek, demonstrate remarkable performance across various reasoning tasks. However, their application in specialized and highly complex domains, particularly medicine, presents unique and significant challenges that often expose the limitations of general-purpose models. Unlike other fields, medical reasoning demands not only a vast breadth of knowledge, encompassing diverse sub-disciplines from diagnostics to pharmacology, but also an unparalleled depth of understanding, requiring nuanced interpretation of symptoms, patient histories, and intricate biological processes. The inherent complexity of medical knowledge, coupled with the dynamic and evolving nature of medical knowledge, means that traditional LLMs, despite extensive pre-training, frequently struggle with the precision, reliability, and contextual understanding critical for clinical applications.
Existing research attempts to enhance the medical performance of LLMs through various methods, broadly categorized into traditional fine-tuning optimization and reflection-based optimization. Traditional fine-tuning approaches, including supervised fine-tuning (SFT) and alignment strategies like reinforcement learning from human feedback (RLHF) or direct preference optimization (DPO), achieve notable progress in general language tasks. However, these methods often optimize for direct answers or simplistic reasoning paths, proving insufficient for the multi-step, intricate reasoning required in complex medical scenarios. The challenges in medicine, such as the need for accurate differential diagnoses, personalized treatment plans, and the integration of multidisciplinary information, necessitate models that can perform deep reasoning rather than merely generating plausible responses.
Recent reflection-based methods aim to overcome these limitations by guiding language models to self-reflect and self-correct during reasoning. While LLMs possess some inherent capacity for self-correction,SLMs  typically lack this ability. Although large models can be employed to correct errors in SLMs, relying on them for this purpose incurs significant computational cost and time consumption, making it an economically unfeasible solution for widespread application. Consequently, existing methods often fail to strike a balance between effectiveness in handling medical complexity and computational efficiency.
To address these specific challenges within the medical domain and overcome the limitations of current LLM training paradigms, we propose MedCritical, a novel two-stage framework designed to significantly improve the reasoning capabilities of SLMs in medical contexts. Our approach is specifically tailored to tackle the dual challenges of broad knowledge scope and deep reasoning requirements in medicine. In the first stage, we employ SFTby extracting high-level and detailed long-chain thought templates from a large teacher model. This process guides the student model to generate more sophisticated and medically relevant reasoning thoughts, thereby addressing the knowledge breadth and initial reasoning depth. In the second stage, we introduce a novel model self-collaboration DPO  mechanism. This stage enhances the reasoning ability of the student model by enabling it to learn from its own error-driven insights through self-iteration and adversarial correction trajectories. This self-learning DPO approach is crucial for consolidating the skills and knowledge of the SLM, allowing it to navigate the intricate and often ambiguous nature of medical problems more effectively and efficiently than traditional knowledge distillation methods. By enabling the model to identify and correct its own weaknesses, MedCritical provides a robust solution for developing highly capable medical SLMs at a significantly lower cost.
Furthermore, we construct high-quality fine-tuning and optimization datasets specifically curated for medical question-answering, comprising 13,000 samples from an original dataset, with a thought-level error correction optimization training set of 3,000 samples. This training set includes medical questions, pre\hyp{}designed thought chain formats, and teacher-generated steps based on real answers, alongside a 10,000-sample test set.
Our contributions are summarized as follows: 

(i) We propose MedCritical, a novel two-stage fine-tuning method specifically designed to enhance the medical reasoning ability of SLMs by addressing the unique challenges of the domain. 

(ii) We introduce model self-collaboration DPO, an innovative approach that leverages a fine-tuned small LLM for self-adversarial learning, thereby significantly improving its reasoning capabilities and overcoming inherent reasoning bottlenecks in complex medical tasks.

(iii) We construct two high-quality, medically-focused datasets and develop a powerful medical reasoning model, achieving an accuracy of 70.62\% on the CMExam dataset, setting a new SOTA performance record among all 7B models.
\section{Related Work}
\label{sec:related_work}
\textbf{Application of Reinforcement Learning with Human Feedback in LLMs} To enhance the performance and reliability of LLMs , the RLHF method proposed by \cite{christiano2017deep} and \cite{ouyang2022training} is introduced for LLM alignment. This method imposes higher requirements on the dataset, as it necessitates paired annotated data to train the reward model, thereby reflecting human preferences. Subsequently, reinforcement learning is employed to train the policy model to maximize the estimated reward. Although this method has proven effective, the process is complex and computationally intensive due to its reliance on the quality of the reward model. To simplify this process, DPO  \cite{rafailov2024direct} is proposed, which directly uses paired data for optimization. By defining the preference loss as a function of the policy, DPO can optimize the policy using simple training techniques, avoiding the complexity of reinforcement learning. However, current methods have limited improvements in medical reasoning, stemming from the design of the optimization unit. Studies such as Step-DPO \cite{lai2024self} establish more granular reward units by treating each intermediate reasoning step as a basic unit. However, they fail to fully leverage the self-learning capabilities of fine-tuned models. This paper specifically designs a model self-collaborative reasoning reward to efficiently utilize this aspect.

\textbf{Thinking Extensions for Medical Reasoning} Thinking extensions for medical reasoning primarily focus on pre-designed reasoning structures or templates to enhance LLM medical reasoning capabilities through prompting techniques. Chain-of-Thought (CoT) prompts \cite{wei2022chain} and their variants \cite{kojima2022large,press2023measuring,arora2022ask}, such as from few to many \cite{zhou2022least}, decomposed prompts \cite{khot2022decomposed}, and automatic CoT \cite{zhang2022automatic} — prompting LLMs to decompose complex problems into simpler sub-tasks, systematically solve them, and then summarize the final answer. Innovative methods such as ``Thinking Trees'' \cite{yao2024tree} and ``Thinking Maps'' \cite{besta2024graph} further complicate the field by exploring dynamic, nonlinear reasoning paths to expand the heuristic capabilities of LLMs \cite{chen2023program,ning2023skeleton}. However, these methods face challenges such as increased resource requirements, higher time complexity, reliance on manual prompt design, and optimization for specific task types. The recently proposed BoT \cite{yang2024buffer} adopts a task-agnostic meta-buffer paradigm to efficiently solve problems based on accumulated thinking templates. However, this is a training-free framework that may not fundamentally enhance LLM's reasoning capabilities. This paper proposes a new paradigm that enables LLM to enhance its reasoning ability through its own right-wrong confrontation, thereby breaking through the limitations of LLM's original thinking and expanding the model's ability to solve a wider range of problems.
\section{Method}
\label{sec:method}
\subsection{Overview of two-stage training framework}
\begin{figure}[ht]
    \centering
    \includegraphics[width=\columnwidth]{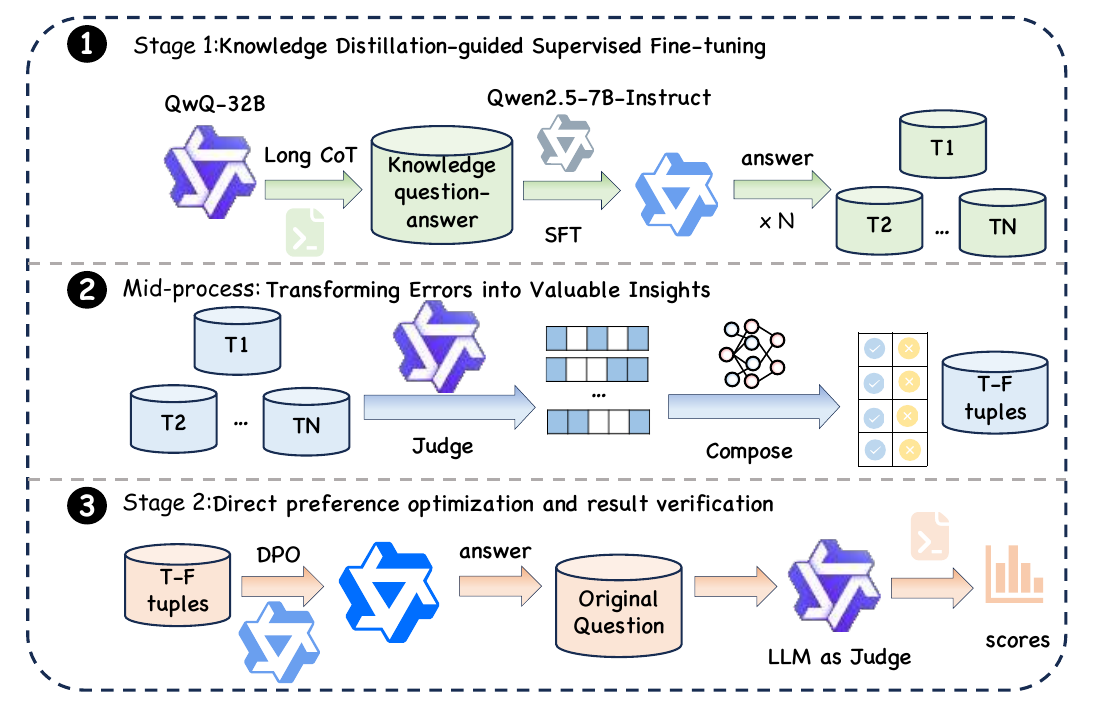}
    \caption{Pipeline of MedCritical framework, illustrating supervised fine-tuning with long chain-of-thought templates and model self-collaborative DPO for enhancing medical reasoning inSLMs.}
    \label{fig:pipeline}
\end{figure}

This section introduces our two-stage training framework in detail and shows our training process in Figure \ref{fig:pipeline}. In the first stage, given a raw medical question-answer set, we use a large language model (see Section 3.2) to generate detailed reasoning chains and answers for it based on a given long thought chain template, and then use these data to fine-tune the small model. In the second stage, based on multiple responses of the fine-tuned small model to the original question, we use the large language model to make correct or incorrect judgments, and combine its correct or incorrect answers to the same question into pairs (see Section 3.3) for DPO to obtain the final result model. The whole process is essentially a self-criticism of the small model after training, so it is named MedCritical.
\subsection{Supervised Fine-Tuning Based on Long Thinking Chain Thinking Templates}
\textbf{Constructing Thinking Templates from Teacher LLMs} \\
Traditional instruction-response datasets used for training LLMs \cite{ouyang2022training} primarily focus on the correctness of responses, leading LLMs to merely simulate given solutions and answers while neglecting the importance of intermediate long thinking chains. Recent studies such as BoT utilize high-level reasoning guidelines (thinking templates) to enable \text{LLMs} to efficiently solve similar problems without additional training. However, for complex and diverse mathematical reasoning tasks, we find that high-level thinking templates alone are insufficient, especially for small \text{LLMs}. To enable small \text{LLMs} to handle complex reasoning tasks, we specifically design a long-chain thinking template extracted from a large teacher \text{LLM} to transfer its capabilities to a small student \text{LLM}. This new long-chain thinking template includes a format requirement specification and a detailed example of long-chain reasoning. To avoid irrelevant information in the template interfering with medical domain problems, the former provides a generic format specification solution for general problems, while the latter details the model's self-reasoning, exploration, and reflection processes in long thinking chains for problems in other domains. Based on this long thinking chain thinking template, we can propose a new fine-tuning objective aimed at integrating human-like problem-solving structures into model reasoning and explicitly generating long thinking chains during the reasoning process. We first collect a set of medical problem instances $\mathcal{D} = \{(\boldsymbol{q}, \boldsymbol{a})\}$, where each instance contains a question $\boldsymbol{q}$ and its direct answer $\boldsymbol{a}$. For each problem $\boldsymbol{q} \in \mathcal{D}$, we first use a predefined prompt $\boldsymbol{P}$ to extract a long reasoning chain $\boldsymbol{c}$ from a teacher state-of-the-art reasoning \text{LLM} (e.g., QwQ-32B). \\
Then, we can obtain a high-quality fine-tuning dataset $\mathcal{D}_{\text{sft}}$:
\begin{equation}
    \mathcal{D}_{\text{sft}} = (\boldsymbol{q}, \boldsymbol{a}, \boldsymbol{c}) \nonumber
\end{equation}
where $\boldsymbol{c}$ is the solution of the long reasoning chain. We provide an example of a long reasoning chain thinking template, as shown in the text box below.
\begin{tcolorbox}[colback=white, colframe=black, title=Long Chain-of-Thought Template Introduction, fontupper=\scriptsize, boxsep=1pt, left=1mm, right=1mm, top=1mm, bottom=1mm, width=\linewidth, halign=left, valign=top, before skip=0pt, after skip=0pt]
\begin{customlist}
    \item[] You are a medical expert. Please structure your response to the question:\{user\_question\}\{question\_mark\} using the format \texttt{<think>}-[here is your chain of thought]-\texttt{</think>}-\texttt{<answer>}-[here is your formal answer]-\texttt{</answer>}.
    \item[] \hspace{1.15em}- \textcolor{rolecolor}{Role: This section defines the model's role as a medical expert and specifies the required response structure (\texttt{<think>} and \texttt{</think>}, \texttt{<answer>} and \texttt{</answer>} tags). It ensures a professional perspective and clear output organization.}
    \item[] Please provide a detailed reasoning process, ensuring the chain of thought is clear, logically rigorous, and progresses step-by-step. Analyze the problem thoroughly, explore possible answers, correct any errors, and ultimately arrive at a conclusion. This is a multi-line text to test the effect on subsequent lines.
    \item[] \hspace{1.15em}- \textcolor{rolecolor}{Role: This section guides the model to use a detailed, step-by-step reasoning process, enhancing medical problem-solving quality.}
    \item[] \texttt{<think>}
Hmm, the user is asking why people on the internet often compare opponents to something worse than Nazis, even making Nazis seem good by comparison. This is a complex question, and I need to clarify the possible aspects first. Initially, the user might notice extreme online rhetoric, where one side uses Nazis to denigrate the other, leading to phrases like ``worse than Nazis,'' which relativizes the original symbol of extreme evil. This could involve several factors: the exaggeration of online speech, lack of historical knowledge, a slippery slope in moral comparisons, and the trend toward polarized communication.
    \texttt{</think>}
    \item[] \hspace{1.15em}- \textcolor{rolecolor}{Role: This section provides a non-medical example of the \texttt{<think>} format, using a detailed, step-by-step reasoning process about internet rhetoric. Its purpose is to guide the model to emulate this depth for medical questions, avoiding interference by focusing on an unrelated domain.}
    \item[] \texttt{<answer>}
The phenomenon of people on the internet labeling opponents as ``worse than Nazis'' and even making Nazis seem glorified is the result of multiple complex social, psychological, and communicative factors. The main causes and logical chain are as follows: \ldots II. Influence of Internet Propagation Mechanisms Emotion-Driven Algorithm Logic Social media platforms favor promoting controversial, emotional content. Extreme rhetoric (e.g., comparing opponents to Nazis) triggers emotions like anger or fear, gaining more visibility. \ldots IV. Consequences and Reflection Undermining Historical Seriousness The uniqueness of Nazi crimes is blurred, weakening public vigilance against true extremism. \ldots Conclusion This phenomenon is a microcosm of the internet era's struggle for discourse dominance: when extremism becomes a shortcut to attention, the complexity of history and reality is sacrificed.
    \texttt{</answer>}
    \item[] \hspace{1.15em}- \textcolor{rolecolor}{Role: This section demonstrates the \texttt{<answer>} format with a non-medical example, summarizing reasoning into a structured conclusion. It guides the model to present concise medical diagnoses or treatments, maintaining clarity without domain overlap.}
    \item[] This question's answer is: \{reference\_answer\}
    \item[] \hspace{1.15em}- \textcolor{rolecolor}{Role: This section subtly embeds the correct answer for medical questions, ensuring the model derives it through reasoning without direct reference, preserving the authenticity of the process.}
    \item[] Please strictly follow the format, starting with the \texttt{<think>} tag for a detailed self-reflection, exploration, and self-correction long chain of thought, ending with \texttt{</think>}. Then begin the formal answer with \texttt{<answer>}, ending with \texttt{</answer>}. The question to address is: \{user\_question\}\{question\_mark\}
    \item[] \hspace{1.15em}- \textcolor{rolecolor}{Role: This section enforces the structured format, ensuring disciplined medical reasoning and response alignment.}
\end{customlist}
\end{tcolorbox}
\BlankLine
\textbf{Thought-based supervised fine-tuning} \\
After organizing the long chain-of-thought dataset $\mathcal{D}_{\text{sft}}$, our optimization goal is to enable student \text{LLMs} $\boldsymbol{\pi}$ to reason using long chains of thought and gain a more comprehensive understanding of each problem-solving process, which can be expressed as:
\begin{equation}
    L_{\text{sft}}(\boldsymbol{\pi}) = \mathbb{E}_{(\boldsymbol{q}, \boldsymbol{a}, \boldsymbol{c}) \sim \mathcal{D}_{\text{sft}}} \left[ \log \boldsymbol{\pi}(\boldsymbol{c}, \boldsymbol{a} | \boldsymbol{q}, \boldsymbol{P}_{\text{stu}}) \right] \nonumber
\end{equation}
Starting from the base student \text{LLM} $\boldsymbol{\pi}$, $L_{\text{sft}}$ achieves optimization by maximizing the likelihood of the response given the prompt $\boldsymbol{P}_{\text{stu}}$ and the input question $\boldsymbol{q}$, where $\boldsymbol{P}_{\text{stu}}$ represents the predefined prompt $\boldsymbol{P}_{\text{tea}}$. After the fine-tuning process, we significantly enhance the reasoning capabilities of the base student \text{LLM} by learning the long chain-of-thought reasoning of state-of-the-art reasoning \text{LLMs}, enabling the student \text{LLM} to generate reasoning similar to that accompanying the final answer. Subsequently, we obtain the fine-tuned student \text{LLM} $\boldsymbol{\pi}_{\text{ref}}$, which can be used for cross-model collaborative DPO in Section 4.2.
\subsection{Model self-improvement DPO}
\textbf{Improving DPO by self-comparison} \\
Although error correction methods dominated by LLMs have proven effective in DPO practice, their optimization objectives are inefficient for complex medical reasoning tasks. As pointed out by \cite{lai2024self}, the problem arises because the monopoly of LLMs makes us ignore the valuable resources generated by small models in large corpora. This may lead to a longer training process without significant improvement in results, because the previously fine-tuned small model's ability to solve is comparable to that of LLMs in the medical field. In addition, it is difficult for \text{LLM} to locate errors and obtain improvements from them \cite{tyen2024error}. This is similar to a student trying to gain subtle insights from a correct long solution. The root of the problem is that there are too many distractions and it is difficult to focus on pointing out the problem and giving the correct solution. To solve this problem, we avoid the overly fine error pointing of LLMs and instead focus on the overall optimization objective, giving priority to instance-level preferences. Specifically, we first let the large language model judge whether the student model is correct, and then use the overall comparison results of correct and incorrect answers in multiple answers to the same question as the optimization unit. This method prioritizes the model self-control trajectory of the student \text{LLM} $\boldsymbol{\pi}_{\text{tea}}$ rather than the minor correction trajectory of the teacher \text{LLM} $\boldsymbol{\pi}_{\text{ref}}$, thereby improving the ability of the student \text{LLM}.

\textbf{Collecting correct and incorrect thinking} \\
In order to achieve correction at the thinking level, we need to collect a dataset containing paired data of correct and incorrect thinking. Specifically, we use the fine-tuned student model $\boldsymbol{\pi}_{\text{ref}}$ to perform multiple reasoning based on thinking chains on the sampled test dataset $\mathcal{D}_{\text{test}} = \{ \boldsymbol{q}_{\text{test}} \}$, and obtain the test result $\mathcal{D}_{\text{pred}} = \{ \boldsymbol{q}_{\text{test}}, \boldsymbol{a}_{\text{test}} \}$. The teacher \text{LLM} makes a binary judgment of correct or incorrect for each $\boldsymbol{a}_{\text{test}}$, and finally obtains the correct-incorrect dataset:
\begin{equation}
\mathcal{D}_{\text{cor-err}} = \{ \boldsymbol{q}_{\text{test}}, \{ \boldsymbol{A}_{\text{test}}^+, \boldsymbol{A}_{\text{test}}^- \} \}, \nonumber
\end{equation}
Where $\boldsymbol{A}_{\text{test}}^+$ is the correct solution set, $\boldsymbol{A}_{\text{test}}^-$ is the incorrect solution set of the corresponding problem. Here, we use powerful models in the current medical reasoning field (such as QwQ-32B) as experienced teacher models $\boldsymbol{\pi}_{\text{tea}}$. To ensure the accuracy of the correct or incorrect binary classification, we design a prompt $\boldsymbol{P}_C$ to guide $\boldsymbol{\pi}_{\text{tea}}$ to deterministically output 1 or 2 (representing correct or incorrect) in the provided answer.

\textbf{Improving Reasoning Ability by Model Self-correction} \\
In the second stage of the method, our proposed model self-collaborative DPO uses the self-opposition of the student \text{LLM} to improve its reasoning and generalization ability. As shown in formula (7), comparing the above correct reasoning $\boldsymbol{A}_{\text{test}}^+$ with the incorrect reasoning $\boldsymbol{A}_{\text{test}}^-$, our model self-collaborative DPO aims to maximize the probability that the student \text{LLM} correctly answers the question $\boldsymbol{q}$. The optimization objective of our model self-cooperative DPO can be expressed as:
\begin{equation}
\begin{split}
L_{\text{DPO-self}}(\theta) = & -\mathbb{E}_{(\boldsymbol{q}, \boldsymbol{A}_{\text{test}}^+, \boldsymbol{A}_{\text{test}}^-) \sim \mathcal{D}_{\text{cor-err}}} \left[ \log \sigma \left( \beta \log \frac{\boldsymbol{\pi}_\theta(\boldsymbol{A}_{\text{test}}^+ | \boldsymbol{q})}{\boldsymbol{\pi}_{\text{ref}}(\boldsymbol{A}_{\text{test}}^+ | \boldsymbol{q})} \right. \right. \\
& \left. \left. - \beta \log \frac{\boldsymbol{\pi}_\theta(\boldsymbol{A}_{\text{test}}^- | \boldsymbol{q})}{\boldsymbol{\pi}_{\text{ref}}(\boldsymbol{A}_{\text{test}}^- | \boldsymbol{q})} \right) \right], \nonumber
\end{split}
\end{equation}
\begin{algorithm}[ht]
\scriptsize
\caption{Pseudocode of MedCritical Framework for Student Model}
\label{alg:critical}
\KwIn{Medical Dataset $\mathcal{D}$, Teacher Model $\boldsymbol{\pi}_{\text{tea}}$, Student Model $\boldsymbol{\pi}_{\text{stu}}$}
\KwOut{Fine-tuned Student Model $\boldsymbol{\pi}_{\text{stu}}$}
\BlankLine
\textbf{Stage 1: Supervised Fine-Tuning with Long Chain-of-Thought} \\
Initialize fine-tuning dataset $\mathcal{D}_{\text{sft}} = \emptyset$\;
\ForEach{$(\boldsymbol{q}, \boldsymbol{a}) \in \mathcal{D}$}{
    $\boldsymbol{c} \gets \boldsymbol{\pi}_{\text{tea}}(\boldsymbol{q}, \boldsymbol{P}_{\text{tea}})$
    Add $(\boldsymbol{q}, \boldsymbol{a}, \boldsymbol{c})$ to $\mathcal{D}_{\text{sft}}$\;
}
\ForEach{$(\boldsymbol{q}, \boldsymbol{a}, \boldsymbol{c}) \in \mathcal{D}_{\text{sft}}$}{
    Update $\boldsymbol{\pi}_{\text{stu}}$ to maximize $\log \boldsymbol{\pi}_{\text{stu}}(\boldsymbol{c}, \boldsymbol{a} | \boldsymbol{q}, \boldsymbol{P}_{\text{stu}})$
}
Set $\boldsymbol{\pi}_{\text{ref}} \gets \boldsymbol{\pi}_{\text{stu}}$
\textbf{Stage 2: Model Self-Collaborative DPO} \\
Initialize correct-erroneous dataset $\mathcal{D}_{\text{cor-err}} = \emptyset$\;
\ForEach{$\boldsymbol{q} \in \mathcal{D}_{\text{test}}$}{
    Initialize $\boldsymbol{A}_{\text{test}}^+ = \emptyset$, $\boldsymbol{A}_{\text{test}}^- = \emptyset$\;
    \For{$i = 1$ to $n$}{
        $\boldsymbol{a}_{\text{test}} \gets \boldsymbol{\pi}_{\text{ref}}(\boldsymbol{q})$
        $j \gets \boldsymbol{\pi}_{\text{tea}}(\boldsymbol{q}, \boldsymbol{a}_{\text{test}}, \boldsymbol{P}_C)$
        \If{$j = 1$}{
            Add $\boldsymbol{a}_{\text{test}}$ to $\boldsymbol{A}_{\text{test}}^+$\;
        }
        \Else{
            Add $\boldsymbol{a}_{\text{test}}$ to $\boldsymbol{A}_{\text{test}}^-$\;
        }
    }
    Add $(\boldsymbol{q}, \boldsymbol{A}_{\text{test}}^+, \boldsymbol{A}_{\text{test}}^-)$ to $\mathcal{D}_{\text{cor-err}}$\;
}
\ForEach{$(\boldsymbol{q}, \boldsymbol{A}_{\text{test}}^+, \boldsymbol{A}_{\text{test}}^-) \in \mathcal{D}_{\text{cor-err}}$}{
    Update $\boldsymbol{\pi}_{\text{stu}}$ to minimize $L_{\text{DPO-self}}(\theta)$
}
\textbf{Output:} Return the fine-tuned Student Model $\boldsymbol{\pi}_{\text{stu}}$\;
\Return{$\boldsymbol{\pi}_{\text{stu}}$}
\end{algorithm}
\section{Experiments}
\label{sec:experiments}
\subsection{Experimental Setup and Baselines}
To evaluate our MedCritical framework, we utilize the CMExam dataset from \url{https://github.com/williamliujl/CMExam}, which originally contains 68,000 samples. We filter 13,887 eligible samples, with a training set of 3,000 for fine-tuning and optimization, and a test set of 10,887 for evaluation. As detailed below, we explain the implementation details of the ten evaluation methods. The baseline Student model is \textit{Qwen2.5-7B-Instruct} and the Teacher model is \textit{QwQ-32B}, which serves as the provider of high-quality reasoning chains. \textit{RawSFT} adopts the standard knowledge distillation method to SFT of the student model using the answers generated by the teacher. \textit{ThoughtSFT} requires the teacher model to generate knowledge chains. \textit{ThoughtDPO}, \textit{AnswerDPO}, and \textit{ExpandDPO} are teacher-guided DPO  methods applied to \textit{ThoughtSFT}. Specifically, \textit{ThoughtDPO} adopts DPO, where the teacher generates correct answers for incorrect answers without providing the original answers; \textit{AnswerDPO} provides the correct answers to the teacher; and \textit{ExpandDPO} augments the dataset by generating three teacher answers for each incorrect answer. In contrast, \textit{SelfDPO} employs a self-improvement mechanism, where \textit{ThoughtSFT} generates four answers for each question, which are judged as correct-wrong answer pairs, all of which are used for DPO training on \textit{SelfDPO}. \textit{TemplateSFT} is trained by introducing Chain of Thoughts (CoT) and example templates derived from the teacher model in the SFT process. Finally, \textit{MedCritical} is similar to \textit{SelfDPO}, but uses \textit{TemplateSFT}, which leverages CoT templates, to generate four answers for improving self-collaborative DPO.
\subsection{Inter-Model Comparison and Analysis}
As shown in Table~\ref{table:performance_results_exp2}, our method achieves new SOTA performance among all 7B models on the CMExam benchmark. It significantly outperforms the strong \textit{Taiyi} by 4.16\% and \textit{Huatuo-o1-7B} by 11.24\%. Notably, our best self-learning model \textit{MedCritical} even achieves superior results than \textit{QwQ-32B} in this test. This result establishes the \textit{MedCritical} model as the leading performer in 7B-level complex medical reasoning tasks. It demonstrates the effectiveness of our proposed \textit{MedCritical} framework in enhancing reasoning capabilities. We attribute the improvement in reasoning accuracy to two aspects. The first CSFT stage provides LLM student models with a deeper and more fine-grained reasoning process. Compared with the traditional reasoning process, it helps LLM student models reason more meticulously, thereby improving the consistency of reasoning and reducing the student model's illusion of ``misconception of its own capabilities''. The second model self-DPO stage helps LLM student models break through the reasoning limitations by comparing the model's own correctness and errors and trying to foster deeper reasoning paths, thereby helping LLM student models gain skills and knowledge to solve problems encountered in partial problem solving. These ``partially or inconsistently solved'' problems are the most important part that students need to consolidate and improve.
\begin{table}[ht]
\centering
\small
\setlength{\tabcolsep}{4pt}
\begin{tabular}{lcccc}
\toprule
\raisebox{-1.2ex}[0pt][0pt]{ModelName} & \raisebox{-1.2ex}[0pt][0pt]{BaseModel} & \multicolumn{3}{c}{Test Sets} \\
\cmidrule(lr){3-5}
 & & TRAIN & TEST & total \\
\midrule
\textbf{\textit{MedCritical}} & \textit{TemplateSFT} & \textbf{68.17} & \textbf{71.29} & \textbf{70.62} \\
\hdashline
\textit{Taiyi} & \textit{Qwen-7B} & 65.03 & 66.86 & 66.46 \\
\textit{Huatuo-o1-7B} & \textit{Qwen2.5-7B} & 54.43 & 60.75 & 59.38 \\
\textit{DeepSeek-R1-Distill-Qwen-7B} & \textit{Qwen2.5-Math-7B} & 56.00 & 57.87 & 57.47 \\
\textit{Lingshu-7B} & \textit{Qwen2.5-VL-7B-Instruct} & 46.40 & 48.89 & 48.35 \\
\textit{Baichuan2-7B-Chat} & \textit{Baichuan2-7B-Base} & 22.43 & 21.81 & 21.94 \\
\textit{Apollo2-7B} & \textit{Qwen2-7B} & 9.60 & 13.70 & 12.81 \\
\bottomrule
\end{tabular}
\vspace{5pt}
\caption{Performance comparison of \textit{MedCritical} with other 7B models on the CMExam dataset, demonstrating superior medical reasoning accuracy and state-of-the-art performance at 70.62\%.}
\label{table:performance_results_exp2}
\end{table}
\begin{figure}[ht]
    \centering
    \includegraphics[width=\columnwidth]{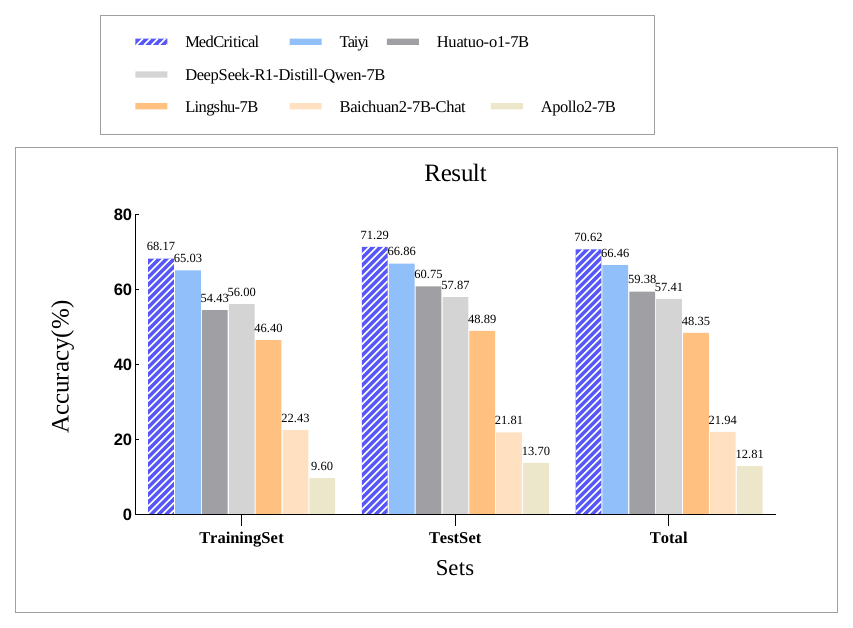} 
    \caption{Cross-model comparison of \textit{MedCritical} against other 7B models on CMExam dataset, showcasing superior medical reasoning accuracy and state-of-the-art performance at 70.62\% in complex reasoning tasks.}
    \label{fig:results_2}
\end{figure}
\subsection{Intra-Model Ablation and Analysis}
We conduct an ablation study on the \textit{MedCritical} model and list the results in Table~\ref{table:performance_results_exp1}. The results show that compared with our CSFT model \textit{ThoughtSFT}, the improvement of traditional SFT is limited, and the accuracy rate lags behind by 10.06\%. This demonstrates the effectiveness of the first stage. Based on our CSFT model \textit{ThoughtSFT}, we further apply some DPO methods to compare with one of our Self-DPO models, \textit{SelfDPO}. The results show that our method achieves higher accuracy than the \textit{ThoughtDPO} model by 0.75\%. This underscores the effectiveness of the second stage. To better understand our two-stage method, consider the example shown in the figure. The base model misunderstands the characteristics of ``dry pneumococcal meningitis'' because it does not consider the effect of reduced mucus in the lung membranes implied by dry pneumococcal meningitis. After applying our long chain thinking reasoning template (denoted as CTP) for training, we can see that the model identifies this factor. However, it fails to correctly apply the impact of this fact on the patient's lung pain to the model, which is precisely caused by the hallucination problem. Finally, our combined one-two-stage model, \textit{MedCritical}, successfully addresses this issue by accurately understanding dry pneumococcal meningitis.
\vspace{2mm}
\BlankLine
\begin{tcolorbox}[colback=white, colframe=black, title=Comparison of answers from three models, fontupper=\scriptsize, boxsep=1pt, left=1mm, right=1mm, top=1mm, bottom=1mm, width=\linewidth, halign=left, valign=top, before skip=0pt, after skip=0pt]
\begin{customlist}
    \item[] What are the chest pain characteristics of dry pleurisy?
    \item[] \textbf{\textit{Student}}: The chest pain associated with dry pleurisy is typically sharp, localized, and can be quite intense. It often feels like a stabbing sensation and is usually felt on one side of the chest, corresponding to the inflamed area of the pleura. This discomfort can be persistent and may significantly impact a person's ability to move or take deep breaths comfortably. While the pain itself is a primary symptom, \textcolor{errorcolor}{it can also be accompanied by other general symptoms such as a dry cough, mild fever, or a general feeling of malaise}, making the overall experience quite uncomfortable for the individual.
    \item[] \hspace{1.15em}- \textcolor{rolecolor}{Critique: This section presents the \textit{Student} model's response, which describes general symptoms but misses key pain patterns, indicating its limitation in medical reasoning.}
    \item[] \textbf{\textit{TemplateSFT}}: Chest pain stemming from dry pleurisy is characteristically sharp or stabbing in nature, often described as a ``knife-like'' sensation. A key feature of this pain is \textcolor{correctcolor}{exacerbation with respiratory movements}. It tends to significantly worsen when you inhale deeply, cough forcefully, sneeze, or even laugh, as these actions cause the chest wall and lungs to move. Conversely, patients often find that the pain lessens considerably, or may even temporarily disappear, when they hold their breath or take very shallow breaths. This distinct pattern of pain makes it a strong indicator of pleuritic involvement.
    \item[] \hspace{1.15em}- \textcolor{rolecolor}{Critique: This section shows the \textit{TemplateSFT} model's response, correctly noting pain patterns but failing to explain the mechanism, indicating partial improvement.}
    \item[] \textbf{\textit{MedCritical}}: Dry pleurisy results in a distinctive, sharp, and often severe chest pain due to the underlying pathological mechanism. Normally, the two layers of the pleura—the visceral pleura covering the lungs and the parietal pleura lining the chest wall—are smooth and lubricated by a small amount of fluid, allowing them to glide effortlessly past each other during breathing. However, in dry pleurisy, inflammation causes these pleural surfaces to become rough and sticky, and the lubricating fluid may be diminished. \textcolor{correctcolor}{Consequently, with each inhalation and exhalation, these inflamed, roughened surfaces rub directly against one another, creating intense friction. This friction is the direct cause of the sharp pain.} When an individual holds his or her breath, the movement between these inflamed pleural layers is temporarily suspended, thereby eliminating the friction and providing a noticeable, albeit temporary, relief from the agonizing pain. This explains why the pain is so closely linked to the act of breathing.
    \item[] \hspace{1.15em}- \textcolor{rolecolor}{Critique: This section displays the \textit{MedCritical} model's response, offering a detailed and accurate mechanism, demonstrating enhanced reasoning capability.}
\end{customlist}
\end{tcolorbox}
\vspace{3mm}
The performance comparison in Table~\ref{table:performance_results_exp1} yields some key findings. First, the performance of \textit{ThoughtSFT} based on classic CoT is significantly better than that of \textit{RawSFT} and Teacher models, with the overall accuracy increased from 58.47\% and 65.37\% to 68.53\%, which reflects the effectiveness of knowledge distillation. Second, the performance of \textit{TemplateSFT} based on long CoT template is further improved to 69.09\%, which is higher than \textit{RawSFT}, indicating that adding long thinking chain reasoning templates with examples can further enhance the model's reasoning ability. Third, three teacher-guided self-learning policy optimization methods (\textit{ThoughtDPO}, \textit{AnswerDPO}, and \textit{ExpandDPO}) slightly improve over \textit{ThoughtSFT} (68.53\%) (68.75\%, 69.06\%, and 68.75\%), but still lag behind \textit{SelfDPO} (69.50\%), which highlights the advantage of our self-learning method in mining error-driven insights. Finally, \textit{MedCritical} achieves the highest accuracy of 70.62\%, outperforming \textit{SelfDPO} and all other methods, which corroborates the effectiveness of combining long CoT templates with self-collaborative DPO.
We qualitatively analyze the teacher-guided T-DPO model \textit{ExpandDPO} and our Self-DPO model \textit{MedCritical}. For fairness, we choose the same parameter settings and prompting methods for each model. As shown below, both perform more effective analysis based on the same problem. Considering their accuracies (68.75\% vs 70.62\%), our method is more resource-efficient and enables more accurate reasoning compared to T-DPO, thus solving previously unsolvable problems.
\BlankLine
\begin{tcolorbox}[colback=white, colframe=black, title=Comparison of responses from T-DPO and Self-DPO Model, fontupper=\scriptsize, boxsep=1pt, left=1mm, right=1mm, top=1mm, bottom=1mm, width=\linewidth, halign=left, valign=top, before skip=0pt, after skip=0pt]
\begin{customlist}
    \item[] What are the chest pain characteristics of dry pleurisy?
    \item[] \textit{T-DPO}: Dry pleurisy leads to sharp chest pain primarily because the usual lubrication between the two pleural layers is compromised. In a healthy state, a small amount of fluid allows the visceral and parietal pleura to glide smoothly. However, with dry pleurisy, this fluid may be reduced or absent, and the pleural surfaces become inflamed and sticky. Consequently, during inhalation and exhalation, these unlubricated, inflamed membranes come into direct, abrasive contact. This direct rubbing is the source of the severe pain, which is exacerbated by breathing. Holding one's breath stops this abrasive contact, thus alleviating the pain.
    \item[] \textit{Self-DPO}: Dry pleurisy results in a distinctive, sharp, and often severe chest pain due to the underlying pathological mechanism. Normally, the two layers of the pleura—the visceral pleura covering the lungs and the parietal pleura lining the chest wall—are smooth and lubricated by a small amount of fluid, allowing them to glide effortlessly past each other during breathing. However, in dry pleurisy, inflammation causes these pleural surfaces to become rough and sticky, and the lubricating fluid may be diminished. Consequently, with each inhalation and exhalation, these inflamed, roughened surfaces rub directly against one another, creating intense friction. This friction is the direct cause of the sharp pain. When an individual holds his or her breath, the movement between these inflamed pleural layers is temporarily suspended, thereby eliminating the friction and providing a noticeable, albeit temporary, relief from the agonizing pain. This explains why the pain is so closely linked to the act of breathing.
\end{customlist}
\end{tcolorbox}
\begin{table}[ht]
\centering
\small
\setlength{\tabcolsep}{4pt}
\begin{tabular}{lcccc}
\toprule
\raisebox{-1.2ex}[0pt][0pt]{ModelName} & \raisebox{-1.2ex}[0pt][0pt]{BaseModel} & \multicolumn{3}{c}{Sets} \\
\cmidrule(lr){3-5}
 & & Train & Test & Total \\
\midrule
\textit{Student} & \textit{Qwen2.5-7B-Instruct} & 47.57 & 52.81 & 51.68 \\
\textit{Teacher} & \textit{QwQ-32B} & 59.07 & 67.52 & 65.37 \\
\hdashline
\textit{RawSFT} & \textit{Student} & 55.37 & 59.33 & 58.47 \\
\textit{ThoughtSFT} & \textit{Student} & 64.87 & 69.54 & 68.53 \\
\textit{ThoughtDPO} & \textit{ThoughtSFT} & 62.43 & 70.49 & 68.75 \\
\textit{AnswerDPO} & \textit{ThoughtSFT} & 63.63 & 70.56 & 69.06 \\
\textit{ExpandDPO} & \textit{ThoughtSFT} & 64.27 & 69.98 & 68.75 \\
\textit{SelfDPO} & \textit{ThoughtSFT} & 66.27 & 70.39 & 69.50 \\
\textit{TemplateSFT} & \textit{Student} & 66.03 & 69.93 & 69.09 \\
\textbf{\textit{MedCritical}} & \textit{TemplateSFT} & \textbf{68.17} & \textbf{71.29} & \textbf{70.62} \\
\bottomrule
\end{tabular}
\vspace{5pt}
\caption{Performance comparison of student, teacher, and fine-tuned models across training, test, and total sets, evaluating reasoning accuracy in ablation study.}
\label{table:performance_results_exp1}
\end{table}
\begin{figure}[ht]
    \centering
    \includegraphics[width=\columnwidth]{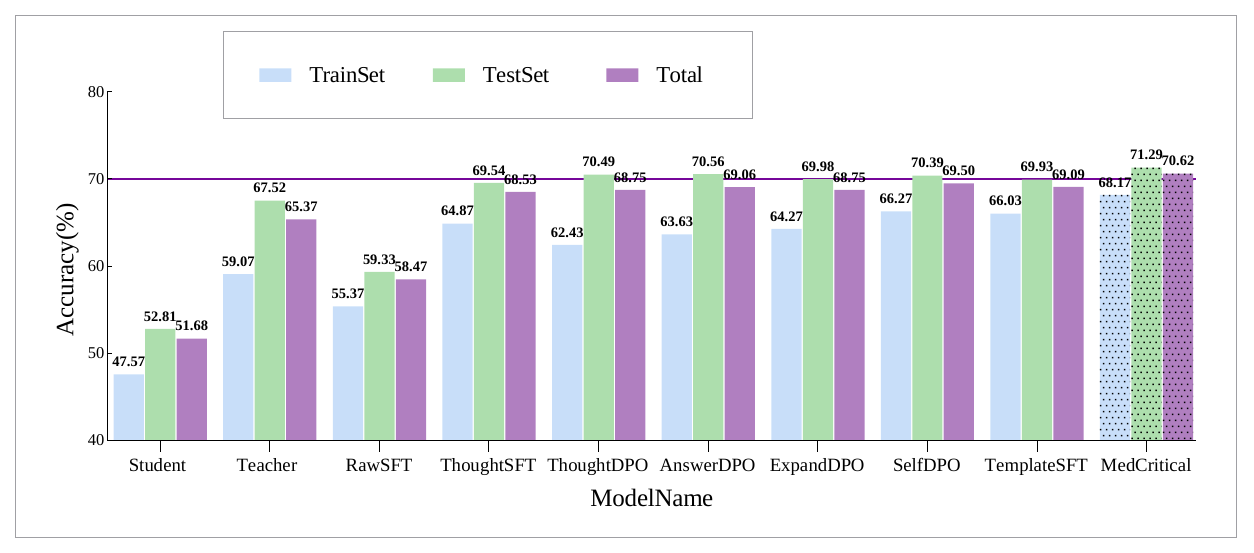}
    \caption{Self-comparison of \textit{MedCritical} model with various fine-tuning and optimization methods, demonstrating improved reasoning accuracy across training and test sets in ablation study for medical reasoning tasks.}
    \label{fig:results_1}
\end{figure}
\section{Conclusion}
\label{sec:conclusion}
This paper proposes a novel two-stage framework that can significantly improve the reasoning ability of language models. We propose fine-tuning based on long thought chain templates to enable \text{LLM} to produce more sophisticated reasoning thinking, and introduce model self-collaborative DPO to improve the reasoning ability of student \text{LLM} through self-correct confrontation on the same problem. Experiments show that our method is superior to previous methods. In the CMExam benchmark, our method is 4.16\% more accurate than the strong \textit{Taiyi} and 11.24\% higher than \textit{Huatuo-o1-7B}. In future work, we will extend this new framework to larger models and more complex datasets.
\section{Acknowledgements}
This research is funded by the FDUROP program of Fudan University (No. 24254) and supported by the CFFF computing platform.
\bibliographystyle{elsarticle-num}
\bibliography{references}
\end{document}